\documentclass[onefignum,onetabnum]{siamart171218} 

\usepackage{lipsum}
\usepackage{amsfonts}
\usepackage{graphicx}
\usepackage{epstopdf}
\usepackage{algorithmic}
\usepackage{textcomp}
\ifpdf
  \DeclareGraphicsExtensions{.eps,.pdf,.png,.jpg}
\else
  \DeclareGraphicsExtensions{.eps}
\fi

\newsiamremark{remark}{Remark}
\newsiamremark{hypothesis}{Hypothesis}
\crefname{hypothesis}{Hypothesis}{Hypotheses}
\newsiamthm{claim}{Claim}

\headers{Locally Linear Attributes of ReLU Neural Networks}{Ben Sattelberg, Renzo Cavalieri, Michael Kirby, Chris Peterson, Ross Beveridge}

\title{Locally Linear Attributes of ReLU Neural Networks}

\author{Ben Sattelberg\thanks{Department of Computer Science, Colorado State University, Fort Collins, Colorado, USA (\email{Ben.Sattelberg@colostate.edu}, \email{Ross.Beveridge@colostate.edu})}
\and Renzo Cavalieri\thanks{Department of Mathematics, Colorado State University, Fort Collins, Colorado, USA (\email{renzo.cavalieri@colostate.edu
}, \email{michael.kirby@colostate.edu
}, \email{peterson@math.colostate.edu})}
\and Michael Kirby\footnotemark[2]
\and Chris Peterson\footnotemark[2]
\and Ross Beveridge\footnotemark[1]
}

\usepackage{amsopn}

\makeatletter
\newcommand*{\addFileDependency}[1]{
  \typeout{(#1)}
  \@addtofilelist{#1}
  \IfFileExists{#1}{}{\typeout{No file #1.}}
}
\makeatother

\ifpdf
\hypersetup{
  pdftitle={Locally Linear Attributes of ReLU Neural Networks},
  pdfauthor={Ben Sattelberg, Renzo Cavalieri, Michael Kirby, Chris Peterson, Ross Beveridge} 
}
\fi
\headers{Locally Linear Attributes of ReLU Neural Networks}{B. Sattelberg, R. Cavalieri, M. Kirby, C. Peterson, R. Beveridge}

\usepackage{multirow}
\usepackage{array}
\usepackage{hyperref}
\usepackage{tikz}
\usetikzlibrary{positioning}
\usepackage{subcaption}

\newcommand{\matr}[1]{\mathbf{#1}}
\newcolumntype{C}[1]{>{\centering\arraybackslash}m{#1}} 
\newcommand{\tikzmark}[2]{\tikz[overlay,remember picture,baseline] \node [anchor=base] (#1) {$#2$};}

\newcommand{\DrawLine}[3][]{%
  \begin{tikzpicture}[overlay,remember picture]
    \draw[#1] (#2.west) -- (#3.east);
  \end{tikzpicture}
}

\def\etal{\emph{et al}.}

\begin{document}

\maketitle

\begin{abstract}
  A ReLU neural network determines/is a continuous piecewise linear map from an input space to an output space. The weights in the neural network determine a decomposition of the input space into convex polytopes and on each of these polytopes the network can be described by a single affine mapping.
  The structure of the decomposition, together with the affine map attached to each polytope, can be analyzed to investigate the behavior of the associated neural network.
\end{abstract}



\section{Introduction}

Building a better understanding of neural network behavior is critically important.  Neural networks are state-of-the-art in a variety of contexts including facial recognition~\cite{arcface} and object recognition~\cite{imagenet}.  However, there is limited understanding of how these networks work or what they are truly doing to achieve such high performance.  We present one path for building understanding and intuition by investigating the locally linear behavior of ReLU networks.

We investigate the linear region facets of ReLU neural networks --- the small regions where the network behaves as a linear function.  These can be considered both through the underlying piecewise linear structure of the network and through the gradient of the network in each region.  Prior work has been done on establishing theoretical bounds on the number of regions that it is possible for a network to have~\cite{montufar2014number, pascanu2013number, raghu2017expressive} and on investigating metrics involving these structures~\cite{novak2018sensitivity}.

We investigate the behavior of these facets for small networks trained on easily visualized problems and on larger, more modern networks trained to recognize handwritten digits~\cite{szegedy2016rethinking, lin2018resnet,MNIST}.  We determine that clustering these facets can be carried out while preserving much of the performance of the networks and that the facets of two different networks, trained on the same problem, are related by a linear map that maintains high accuracy.  In related work done by McNeely-White~\etal~\cite{mcneely2019inception}, it was shown that one can apply a linear map to the feature vector (the outputs of the pre-classification layer) of one network to obtain a vector, considered as a feature vector in the second network, that can then be used by the second network for classification while maintaining high accuracy. The clustering results suggest that networks have significant redundancy at the facet level while the existence of the linear map suggests that networks follow qualitatively similar methods to solve problems.  Although our methods are not currently usable for compression or simplification, they are useful for investigating the behavior of networks.  There are aspects of what is being presented here that may be obvious to those that have thought about neural networks, but we present them to further build intuition for network behavior.

\subsection{Outline}
We first provide an overview of previous work before defining and providing an overview of linear region structures.  We provide illustrated examples of linear regions in different contexts in Section~\ref{sec:linear-regions}.  We then investigate potential applications of these methods:
\begin{itemize}
    \item In Section~\ref{sec:animation}, we show how the linear regions change throughout the training process of simple networks trained on two-dimensional problems, and how the structure of early layers induce behavior in later layers.
    \item In Section~\ref{sec:simplification}, we show experimentally that the number of linear regions a network has is not necessarily a perfect predictor of its complexity, as those linear regions can be clustered while preserving the accuracy of the network.
    \item In Section~\ref{sec:mapping}, we show experimentally that the linear regions of different networks exhibit similarity and that a simple affine mapping between those linear regions maintains a high level of classification accuracy.
\end{itemize}

\section{Related Work}

Much of the original work dealing with the linear regions of ReLU neural networks has focused on investigating expressivity and complexity.  It has previously been shown that networks are universal approximators, that is, subject to certain mild constraints, they are able to approximate any well-behaved function to within arbitrary precision as the size of the network increases~\cite{hanin2017approximating, cybenko1989approximation, hornik1991approximation, lu2017expressive, lin2018resnet}.  As meaningful as these results are, they are typically not applicable to practical neural networks and do not say anything about the expressivity of a \textit{given} neural network.  To assist with determining the expressivity of networks in practice, various groups found and improved bounds on the maximum number of linear regions that ReLU neural networks can have~\cite{montufar2014number, pascanu2013number, raghu2017expressive}.  The main result of this work is that the maximum number of linear regions a network can have grows polynomially in the width and exponentially in the depth~\cite{raghu2017expressive}.  This partially explains the success of the trend in many modern neural networks to go deeper, such as ResNet~\cite{he2016deep}.

However, empirical investigations of the number of linear regions actually achieved by many neural networks have shown different results.  Untrained neural networks after initialization have a number of linear regions that tends to grow linearly in the number of ReLU functions along any one-dimensional subspace of the input space~\cite{hanin2019complexity}.  Furthermore, they tend to grow polynomially in the number of ReLU nodes in the network and exponentially in the dimension of the inputs to the network~\cite{hanin2019deep}.

These linear regions have also been used empirically to measure the sensitivity of neural networks.  As will be discussed in Section~\ref{sec:linear-regions}, the Jacobian of a neural network at a point, together with the value of the neural network at the point, describes exactly the linear function that agrees with the network in a polytope around that point.   Novak~\etal~\cite{novak2018sensitivity} utilized this fact to investigate the effect of hyperparameters on input sensitivity and found that overparameterization can help in generalization.  Additionally, they and Zhang~\etal~\cite{zhang2020empirical} investigated how the linear region structure can be used to predict the quality of a network.

Zhang~\etal~\cite{zhang2018tropical} showed that due to the piecewise linear structure of these neural networks, and under certain assumptions, the set of ReLU neural networks, the set of piecewise linear functions, and the set of tropical rational functions are equivalent.  We do not extend our results to the realm of tropical algebra, but we do take inspiration from the concept of the dual as commonly expressed in tropical algebra.

\section{Linear Regions}
\label{sec:linear-regions}

Neural networks with piecewise linear activation functions, such as ReLU, are continuous piecewise linear maps from the input space to the output space~\cite{zhang2018tropical}.  Additionally, each of the linear portions of this mapping is supported on a convex polytope.

\subsection{Definition}

\begin{figure}[]
\centering
    \begin{tikzpicture}[x=1.5cm, y=1.5cm, >=stealth, node distance = 0.5cm and 1cm, neuron/.style={circle, minimum size=0.75cm}]
    
        \node [neuron,fill=green!50] (input-1) at (0,0) {$x_1$};
        
        \foreach \m [count=\y] in {2,...,3}
            \node [neuron,fill=green!50, below = of input-\y ] (input-\m) {$x_\m$};
         
        \foreach \m [count=\y] in {1,...,3} 
            \node [neuron,fill=red!50, right = of input-\m] (hidden-\m) {};
        
        \foreach \m [count=\y] in {1,...,3}
            \node [neuron,fill=red!50, right = of hidden-\m] (hidden2-\m) {};
            
        \foreach \m [count=\y] in {1,...,3}
            \node [neuron,fill=blue!50, right = of hidden2-\m] (output-\m) {};

        \foreach \m [count=\y] in {2} 
            \draw (hidden-\m |- hidden-\m) node[neuron] {\Huge$\times$};
          
        \foreach \m [count=\y] in {1,3} 
            \draw (hidden2-\m |- hidden2-\m) node[neuron] {\Huge$\times$};

        \node [neuron,fill=gray!50, below=of input-3] (input-4) {1};
        \node [neuron,fill=gray!50, below=of hidden-3] (hidden-4) {1};
        \node [neuron,fill=gray!50, below=of hidden2-3] (hidden2-4) {1};
        
        \foreach \i in {1,...,4}
          \foreach \j in {1,...,3}
            \draw [->] (input-\i) -- (hidden-\j);
            
        \foreach \i in {1,...,4}
          \foreach \j in {1,...,3}
            \draw [->] (hidden-\i) -- (hidden2-\j);
        
        \foreach \i in {1,...,4}
          \foreach \j in {1,...,3}
            \draw [->] (hidden2-\i) -- (output-\j);
    \end{tikzpicture}
    
    Network:
    \footnotesize
    $$
    f(x_1,x_2,x_3) = 
    \begin{bmatrix} 7 & 9 & 6 \\ 
                    4 & 0 & 9 \\ 
                    7 & 3 & 7 
    \end{bmatrix} 
    \text{ReLU}\left( 
    \begin{bmatrix} 2 & 9 & 5 \\ 
                    1 & 3 & 3 \\ 
                    8 & 7 & 0 
    \end{bmatrix} 
    \text{ReLU}\left(
    \begin{bmatrix} 3 & 0 & 4 \\ 
                    5 & 9 & 1 \\ 
                    6 & 1 & 2 
    \end{bmatrix} 
    \begin{bmatrix} x_1 \\ x_2 \\ x_3 \end{bmatrix} + 
    \begin{bmatrix} 7 \\ 2 \\ 9 \end{bmatrix}
    \right) + 
    \begin{bmatrix} 7 \\ 7 \\ 8 \end{bmatrix}
    \right) + 
    \begin{bmatrix} 5 \\ 5 \\ 2 \end{bmatrix}
    $$
    \normalsize
    
    Post ReLU:
    \small
    \begin{align*}
    f(x_1,x_2,x_3) &= 
    \begin{bmatrix} 7 & 9 & 6 \\ 
                    4 & 0 & 9 \\ 
                    7 & 3 & 7 
    \end{bmatrix} 
    \left( 
    \begin{bmatrix} \tikzmark{s2l}{2} & 9 & \tikzmark{s2r}{5} \\ 
                    1 & 3 & 3 \\ 
                    \tikzmark{s3l}{8} & 7 & \tikzmark{s3r}{0} 
    \end{bmatrix} 
    \left(
    \begin{bmatrix} 3 & 0 & 4 \\ 
                    \tikzmark{s1l}{5} & 9 & \tikzmark{s1r}{1} \\ 
                    6 & 1 & 2
    \end{bmatrix} 
    \begin{bmatrix} x_1 \\ x_2 \\ x_3 \end{bmatrix} + 
    \begin{bmatrix} 7 \\ \tikzmark{s1b}{2} \\ 9 \end{bmatrix}
    \right) + 
    \begin{bmatrix} \tikzmark{s2b}{7} \\ 7 \\ \tikzmark{s3b}{8} \end{bmatrix}
    \right) + 
    \begin{bmatrix} 5 \\ 5 \\ 2 \end{bmatrix} 
    \\
    &=
    \begin{bmatrix} 7 & 9 & 6 \\ 
                    4 & 0 & 9 \\ 
                    7 & 3 & 7 
    \end{bmatrix} 
    \left( 
    \begin{bmatrix} 0 & 0 & 0 \\ 
                    1 & 3 & 3 \\ 
                    0 & 0 & 0
    \end{bmatrix} 
    \left(
    \begin{bmatrix} 3 & 0 & 4 \\ 
                    0 & 0 & 0 \\ 
                    6 & 1 & 2
    \end{bmatrix} 
    \begin{bmatrix} x_1 \\ x_2 \\ x_3 \end{bmatrix} + 
    \begin{bmatrix} 7 \\ 0 \\ 9 \end{bmatrix}
    \right) + 
    \begin{bmatrix} 0 \\ 7 \\ 0 \end{bmatrix}
    \right) + 
    \begin{bmatrix} 5 \\ 5 \\ 2 \end{bmatrix} 
    \\
    &= 
    \begin{bmatrix} 42 & 0 & 120 \\
                    20 & 0 & 27 \\
                    336 & 21 & 0
    \end{bmatrix}
    \begin{bmatrix} x_1 \\ x_ 2 \\ x_3 \end{bmatrix}
    +
    \begin{bmatrix} 695 \\ 376 \\ 562 \end{bmatrix}
    \end{align*}
    \normalsize
    \DrawLine[red, very thick, opacity=0.75]{s1l}{s1r}
    \DrawLine[red, very thick, opacity=0.75]{s1b}{s1b}
    \DrawLine[red, very thick, opacity=0.75]{s2l}{s2r}
    \DrawLine[red, very thick, opacity=0.75]{s2b}{s2b}
    \DrawLine[red, very thick, opacity=0.75]{s3l}{s3r}
    \DrawLine[red, very thick, opacity=0.75]{s3b}{s3b}
    
    Region of validity:
    $$
    \begin{aligned}[t]
        \begin{bmatrix} 3 & 0 & 4 \\
                        6 & 1 & 2
        \end{bmatrix}
        \begin{bmatrix} x_1 \\ x_2 \\ x_3 \end{bmatrix} + 
        \begin{bmatrix} 7 \\ 9 \end{bmatrix}
        &\geq \begin{bmatrix} 0 \\ 0 \end{bmatrix}
        &\begin{bmatrix} 1 & 3 & 3 \end{bmatrix} \left(
        \begin{bmatrix} 3 & 0 & 4 \\
                        5 & 9 & 1 \\
                        6 & 1 & 2
        \end{bmatrix}
        \begin{bmatrix} x_1 \\ x_2 \\ x_3 \end{bmatrix} + 
        \begin{bmatrix} 7 \\ 2 \\ 9 \end{bmatrix}
        \right)
        + 7
        &\geq 0\\
        \begin{bmatrix} 5 & 9 & 1 \end{bmatrix} \begin{bmatrix} x_1 \\ x_2 \\ x_3 \end{bmatrix} + 2 &< 0
        &\begin{bmatrix} 2 & 9 & 5 \\
                        8 & 7 & 0
        \end{bmatrix} \left(
        \begin{bmatrix} 3 & 0 & 4 \\
                        5 & 9 & 1 \\
                        6 & 1 & 2
        \end{bmatrix}
        \begin{bmatrix} x_1 \\ x_2 \\ x_3 \end{bmatrix} + 
        \begin{bmatrix} 7 \\ 2 \\ 9 \end{bmatrix}
        \right)
        + \begin{bmatrix} 7 \\ 8 \end{bmatrix}
        &< \begin{bmatrix} 0 \\ 0 \end{bmatrix}
    \end{aligned}
    $$
    
    \caption{An illustration of how the ReLU activation pattern for an input determines the linear mapping used for that input. The region of validity refers to the possible $x$ values for which this ReLU activation pattern exists.  All the equations must be satisfied.}
    \label{fig:linearRegion}
\end{figure}

The piecewise linear and convex polytope structures of a ReLU neural network, $f:\mathbb{R}^d \to \mathbb{R}^o$ with $o$ outputs and inputs in $\mathbb{R}^d$, mean that it can be written as
\begin{equation}
\label{eqn:linear-regions}
    f(x) = \begin{cases} 
            \matr{W}_1 x + b_1, & \text{if } \matr{A}_1 x \leq c_1 \\
            \matr{W}_2 x + b_2, & \text{if } \matr{A}_2 x \leq c_2 \\
            \vdots \\
            \matr{W}_m x + b_m, & \text{if } \matr{A}_m x \leq c_m.
           \end{cases}
\end{equation}

For each $i$, the affine mapping defined by $\matr{W}_i$ and $b_i$ is valid on the convex polytope defined by $\matr{A}_i$ and $c_i$. One can find the values for these parameters, that are valid at $x$, as follows. From a given value of $x$, one finds the associated ReLU activation pattern in the network. From  this activation pattern one determines the affine function, from input space to output  space, that agrees with the neural network at $x$ (i.e. the values of $\matr{W}_i$ and $b_i$). Next, one determines the region of  validity for this linear  function (i.e. the values of $\matr{A}_i$ and $c_i$). Putting it together, $\matr{W}_i$ and $b_i$ determine  the affine linear function that agrees with the ReLU neural network on the polytope, defined by $\matr{A}_i$ and $c_i$, that contains $x$. The process of how one of these $\matr{W}_i$, $b_i$ pairs can be calculated is shown in Figure~\ref{fig:linearRegion}.  This piecewise linear mapping structure can be extended to various other common layers types, such as max and average pooling.

The $\matr{W}_i$ and $\matr{A}_i$ are also linked --- the $\matr{W}_i$ are selected based on which ReLU nodes are activated, and the $\matr{A}_i$ describe where ReLU nodes switch from activated to deactivated or vice-versa. This is partially illustrated in Figure~\ref{fig:linearRegion} and a specific, smaller example of this is shown later in Equations~\ref{eqn:XORnet} and~\ref{eqn:XORlinear}.  There are also similarities and relationships between different $\matr{W}_i$ or $\matr{A}_i$ --- because they are coming from the same network weights with rows removed, there is an inherent structure in the specific values used to construct them.

An additional note to make is that in general, the number of regions, $m$, has the potential to be very large with exponential growth in the depth and polynomial growth in the width of the network~\cite{montufar2014number, pascanu2013number, raghu2017expressive}.  Experimentally, trained networks have been shown to typically exhibit polynomial growth with the number of ReLU activations of the network, where the degree of the polynomial is the input dimension~\cite{hanin2019deep}.  Although this is polynomial, networks applied in domains such as image recognition frequently have inputs with at least 1,000 dimensions, so this still results in very large numbers of regions~\cite{imagenet}.  

The linear mapping network definition, Equation~\ref{eqn:linear-regions}, highlights the fact that as long as one of the ReLU nodes does not switch from ``activated'' to ``deactivated'' or vice-versa, the behavior of the network is purely linear.  Since the network is a composition of continuous linear and piecewise linear functions, it is itself a continuous piecewise linear function that splits the input space into disjoint polytopes, on each of which there is an associated affine mapping. This represents an unequivocally simple way to conceptualize what ReLU networks compute, but unfortunately, the typically extreme growth in the number of facets in  Equation~\ref{eqn:linear-regions} means enumerating the full set of affine mappings is wildly impractical.  

Equation~\ref{eqn:linear-regions} is of conceptual value but arguably by itself not of much practical value, but it leads to several distinct, yet ultimately equivalent views, of neural networks. These views are: 

\begin{itemize}
    \item The weight matrix, $\matr{W}_i$, is the Jacobian of the neural network in the region described by $\matr{A}_i$.  The $j^{\text{th}}$ row of $\matr{W}_i$ is the gradient of the $j^{\text{th}}$ output of the network.  This fact has been utilized previously to consider sensitivity metrics for neural networks~\cite{novak2018sensitivity}.  This also allows for simple calculation of the $\matr{W}_i$ and $b_i$ values.
    \item The weight matrices, $\matr{W}_i$, and biases, $b_i$, form a set of linear maps which the neural network chooses from based on the value of the input.  Each row of these $\matr{W}_i$ is a surface normal to the hyperplane used for classification.
    \item The choices are based on the location of the input in a set of connected polytopes induced by the ReLU structure of the network.  We provide animations showing how these structures evolve as networks train in Section~\ref{sec:animation}.
    \item Each row of $\matr{W}_i$ concatenated with the corresponding element  of $b_i$ forms a point in $\mathbb{R}^{d+1}$.  These points can be considered as lying in a ``dual'' space to the corresponding output of the network, and their structure is analyzed in that context.  We show how this space forms in this section and Section~\ref{sec:animation}, and analyze this space for clustering and similarity of networks in Sections~\ref{sec:simplification} and~\ref{sec:mapping}.
\end{itemize}

\subsection{Example on XOR}

\begin{figure}[h]
    \centering

    \begin{subfigure}[b]{0.45\textwidth}
    \centering
    \renewcommand{\arraystretch}{1.5}
    \begin{tabular}{cc|c}
    $x$ & $y$ & $f(x, y)$ \\ 
    \hline
    -1 & -1 & -1 \\
    1 & -1 & 1 \\
    -1 & 1 & 1 \\
    1 & 1 & -1
    \end{tabular}  
    \caption{\label{fig:XOR-problem}The input and output values of the modified XOR problem.}
    \end{subfigure}
    \hfill
    \begin{subfigure}[b]{0.45\textwidth}
    \centering
    \includegraphics[width=\textwidth]{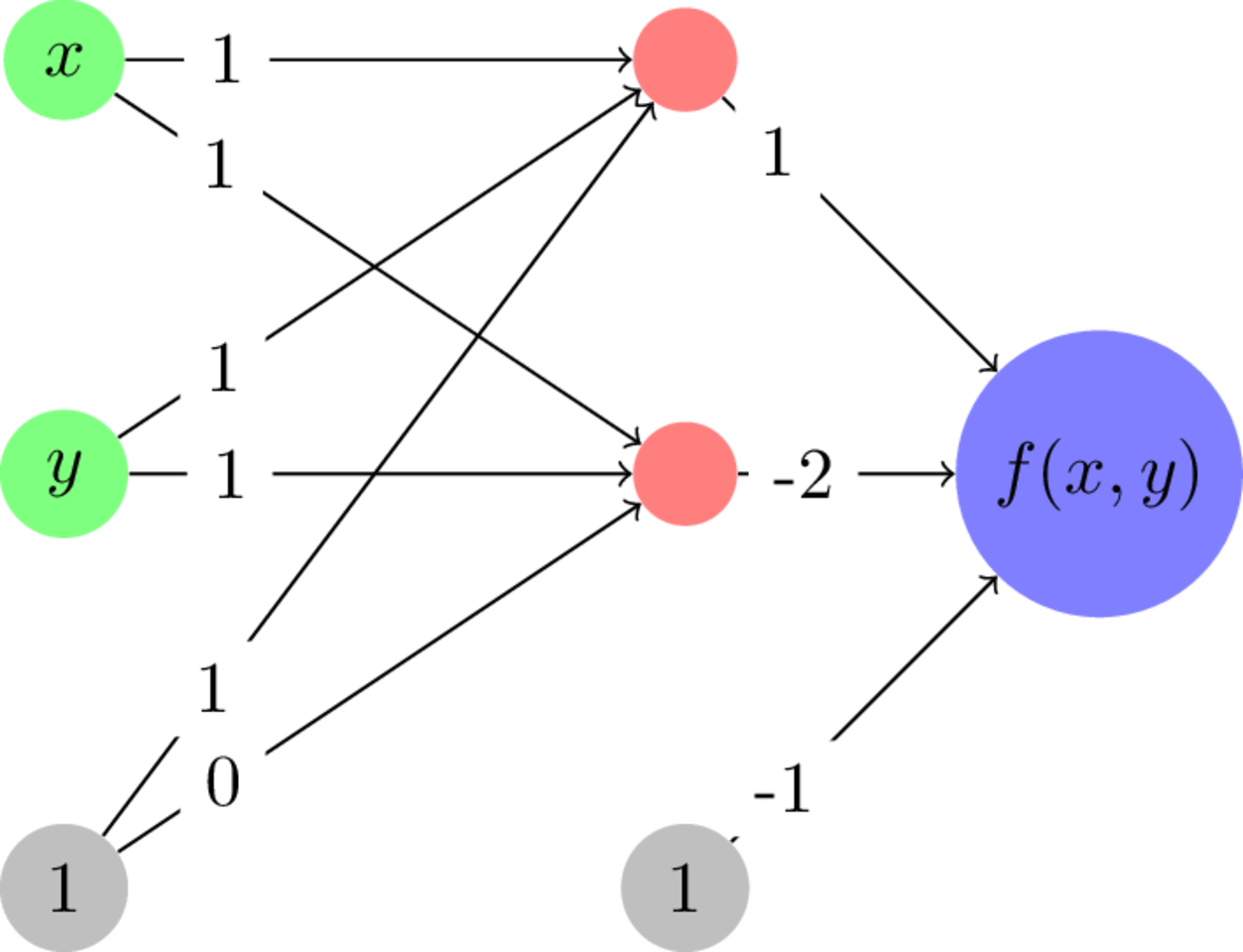}
    \caption{\label{fig:XOR-network-architecture}A network that solves this modified XOR problem.}
    \end{subfigure}
    
    \caption{The modified XOR problem.  (a) the input and output values --- inputs and outputs are rescaled to be from -1 to 1 rather than from 0 to 1.  (b) A network architecture and its associated weights that solves this problem.  Nodes in red have ReLU applied after calculating their associated input values.}
    \label{fig:modxor}
\end{figure}

For an example of how the piecewise linear nature of ReLU neural networks works, we consider the  XOR problem and a ReLU neural network that solves it as presented in Figure~\ref{fig:modxor}. We choose XOR as it is a complex enough problem that it illustrates nonlinear aspects of network behavior, but simple enough that full analysis of that behavior is feasible.  Note that for the XOR function itself, shown in Figure~\ref{fig:XOR-problem}, zero is replaced with minus one to make subsequent examination of network calculations easier. Figure~\ref{fig:XOR-network-architecture} shows a network which solves the XOR problem. The functional form of that same network mapping from the two inputs $x$ and $y$ may be written as
\begin{equation} 
\label{eqn:XORnet}
f(x,y) = \begin{bmatrix} 1 & -2 \end{bmatrix} \max \left\{ \begin{bmatrix} 1 & 1 \\ 1 & 1 \end{bmatrix} \begin{bmatrix} x \\ y \end{bmatrix} + \begin{bmatrix} 1 \\ 0 \end{bmatrix}, \begin{bmatrix} 0 \\ 0 \end{bmatrix} \right\} - 1. 
\end{equation}

As a function on $\mathbb R^2$, the network divides $\mathbb R^2$ into three linear regions with corresponding linear function/polytope pairs,
\begin{equation}
\label{eqn:XORlinear}
f(x) =
        \begin{cases}
        -x-y+1, & \phantom{-}0 \leq x + y, \text{ Both ReLUs activated} \\
        x+y+1, & -1 \leq x + y \leq 0, \text{ Top activated; bottom not activated}\\
        -1, & \phantom{ -1 \leq } x + y \leq -1, \text{ Neither ReLU activated}\\
        \end{cases}
\end{equation}

\begin{figure}[]
    \centering
    \includegraphics[width=0.49\textwidth]{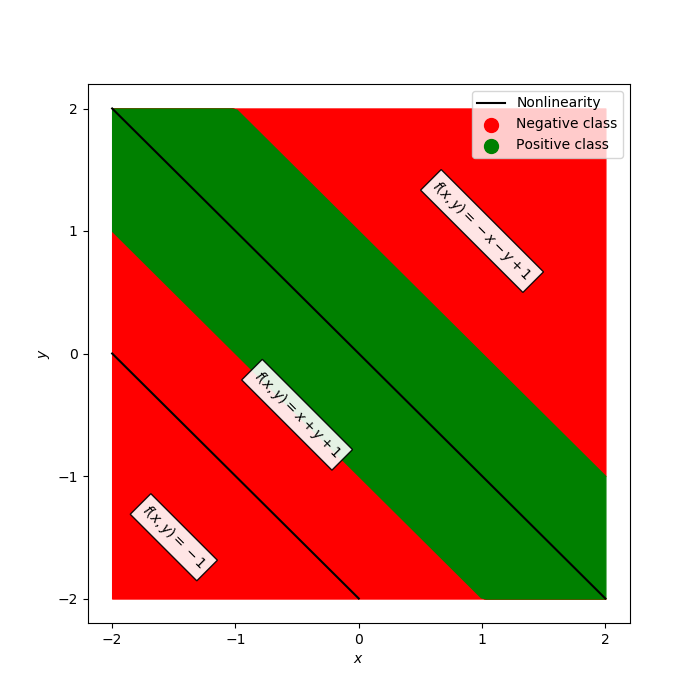}
    \includegraphics[width=0.49\textwidth]{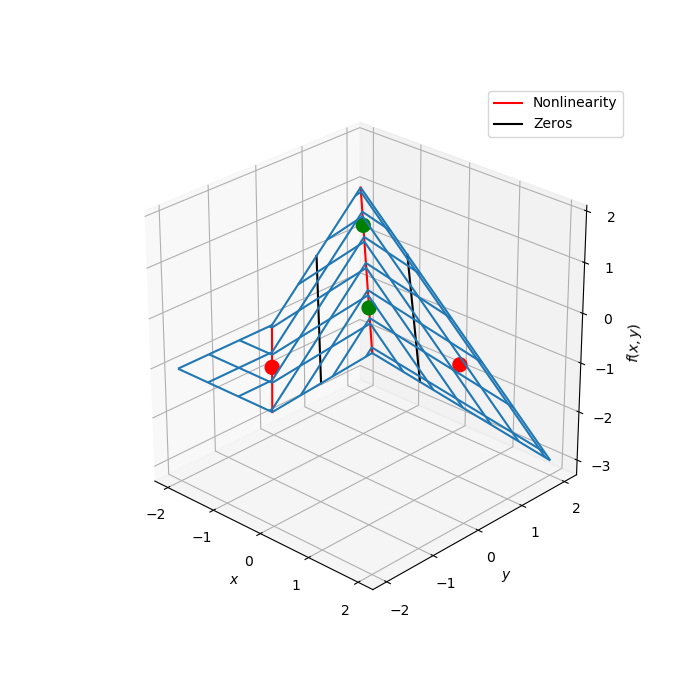}
    \caption{The polytopes and associated linear regions for a simple network to solve the XOR problem.  Left: the cross-section of the network in the plane.  Green corresponds to points that would be labeled in the positive class (neural network output greater than zero) and red corresponds to points that would be labeled in the negative class (neural network output less than zero).  The black lines correspond to the points at which one of the two ReLU units ``activates'' or ``deactivates'' and switches the linear region used for classification.  The three polytopes form bands in the plane.  Right: the surface of the neural network.  The points used for training are shown as green and red dots, the nonlinearities are shown as red lines, and the decision boundary (zeros of the network) are shown as black lines.}
    \label{fig:XORexample}
\end{figure}

These linear regions are shown in Figure~\ref{fig:XORexample}.  Even for this very simple example a complication arises: there is actually a ``fourth'' region, $-4x -4y + 3$, tied to the case where the bottom ReLU unit is activated and the top is not. However, that case occurs in the empty polytope $0 \leq x + y \leq -1$ which cannot occur for any values of $x$ and $y$, and thus in practical terms this empty polytope does not exist. This is an example of a general phenomena where cases exist in principle but are unreachable regardless of input. Further, the existence of such cases explains in part why the number of possible linear regions grows as it does and not simply exponentially in the number of ReLU functions.

There are additional practical complications that can arise but do not on this network due to its simplicity --- a network can be considered as a function on all of $\mathbb R^n$ but the data to which the network is actually applied lies in a bounded region within $\mathbb R^n$.  Polytopes may exist outside of that region but not be meaningful for the given inputs.  Furthermore, in many problems the data used is but a discrete subset of this bounded region. It is possible for the network to define polytopes lying in the bounded region but too small to contain any of the discrete data to which the network is applied.  In general, the number of non-empty polytopes does typically grow beyond the number of actual training samples.

\begin{figure}[]
    \centering
    \includegraphics[width=\textwidth]{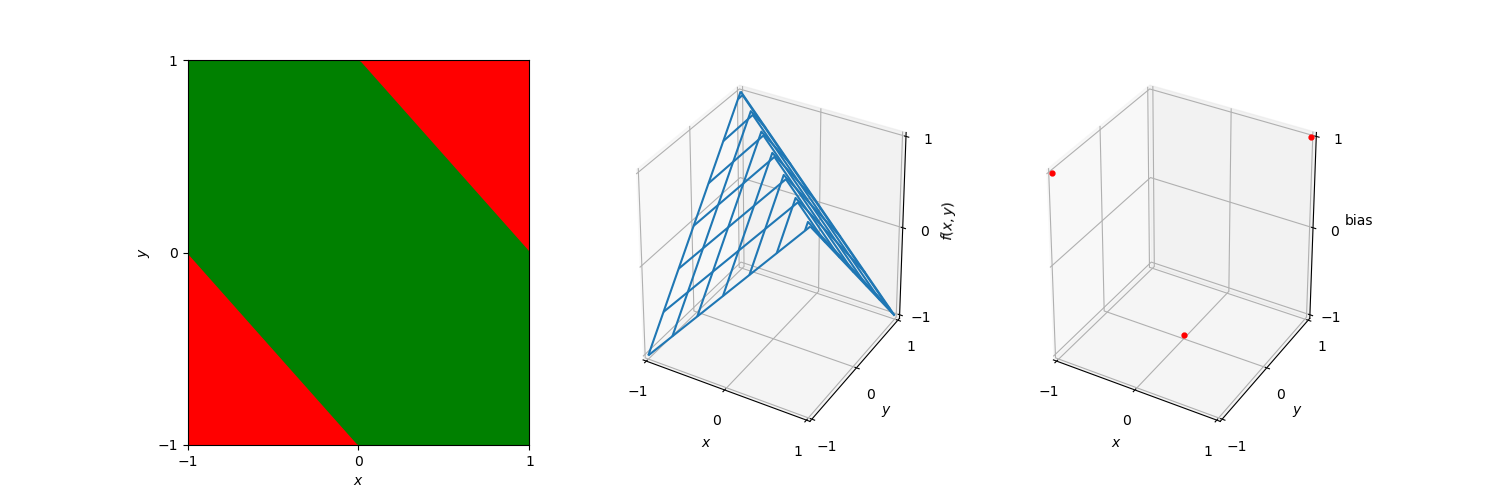}
    
    \includegraphics[width=\textwidth]{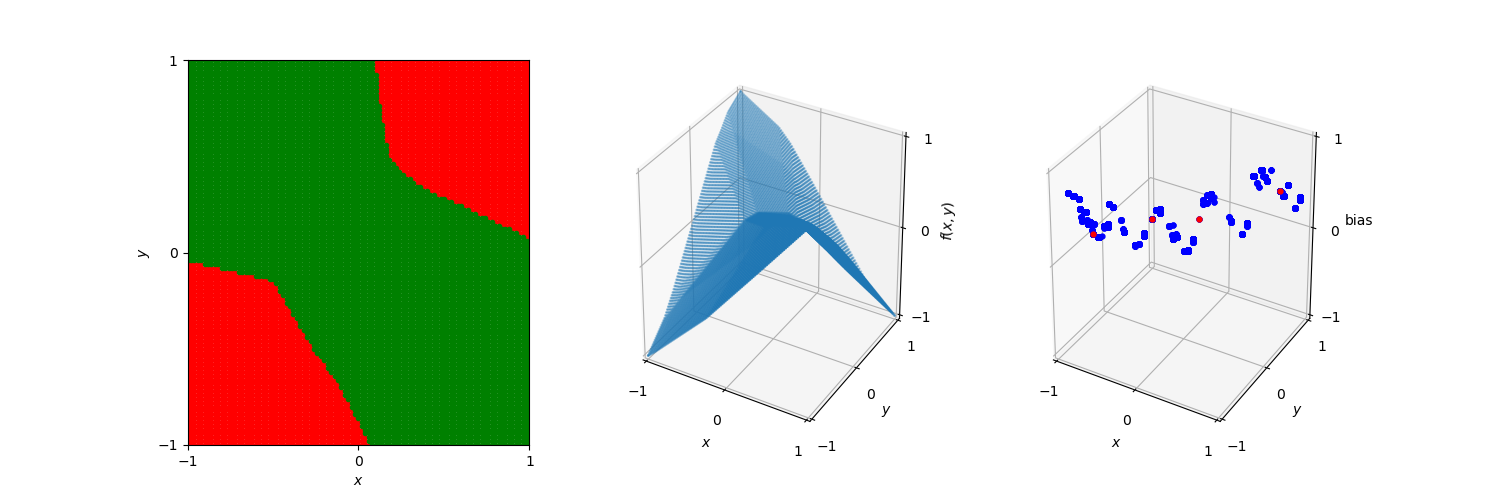}
    
    \includegraphics[width=\textwidth]{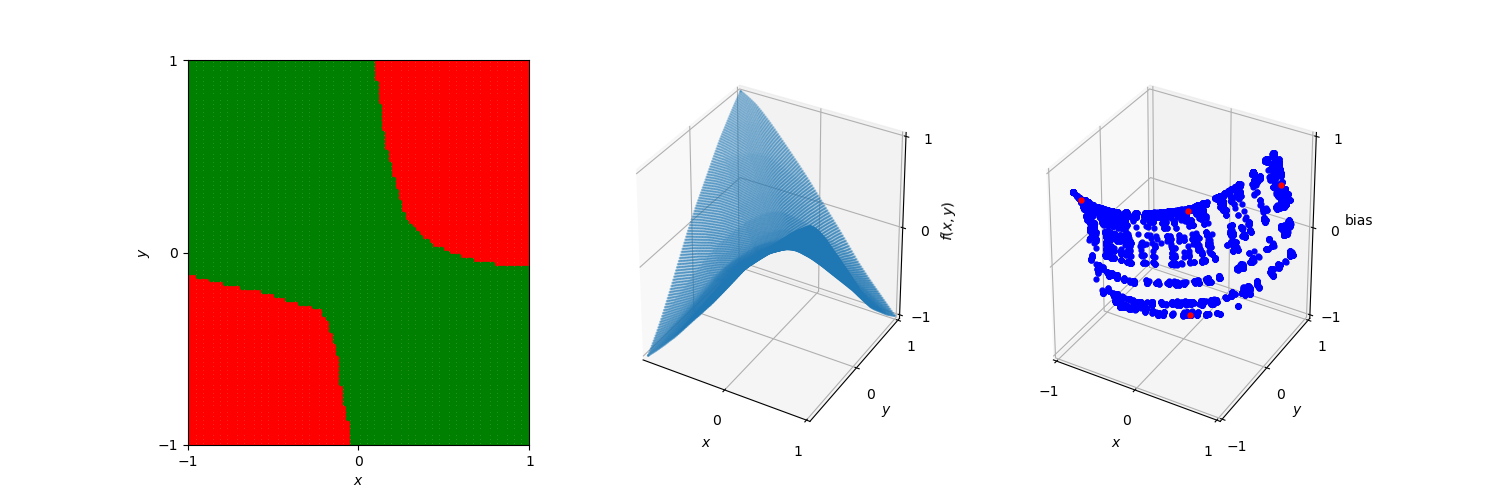}
    \caption{The decision boundaries (left), wireframe representations of output (center), and dual representation of the linear regions (right) for three networks designed to solve ReLU.  The top network is the simple one described previously.  The center and bottom are single hidden layer neural networks with the center having 20 hidden nodes and the bottom having 100 hidden nodes.  In the dual, blue dots represent linear regions used on the $101\times101$ uniform grid in $[-1,1]^2$.  The red dots represent the linear regions used for the actual classification of the four data points --- note that the top image only has three dots corresponding to these, rather than four, as it only has a total of three linear regions.}
    \label{fig:XORduals}
\end{figure}

Returning to the regions shown in Figure~\ref{fig:XORexample}, the weights and biases in these polytopes can be considered as $d+1$-dimensional points existing in a ``dual space'' to the original neural network.  For example,
\begin{equation}
    -x-y+1 = \begin{bmatrix} -1 & -1 & 1 \end{bmatrix} \begin{bmatrix} x \\ y \\ 1 \end{bmatrix}
\end{equation}
and so the point $(-1, -1, 1)$ is induced by this region.  Further examples of these duals are illustrated in Figure~\ref{fig:XORduals}.  These can illustrate patterns in the behavior of the network, and as will be discussed in more detail, mapping between networks or clustering in this space can identify similarity metrics and areas where the neural network gives potentially unnecessary complexity.

\subsection{Polytope Visualization}
\label{sec:polytope-structure}

\begin{figure}[]
    \centering
    \includegraphics[width=0.75\textwidth]{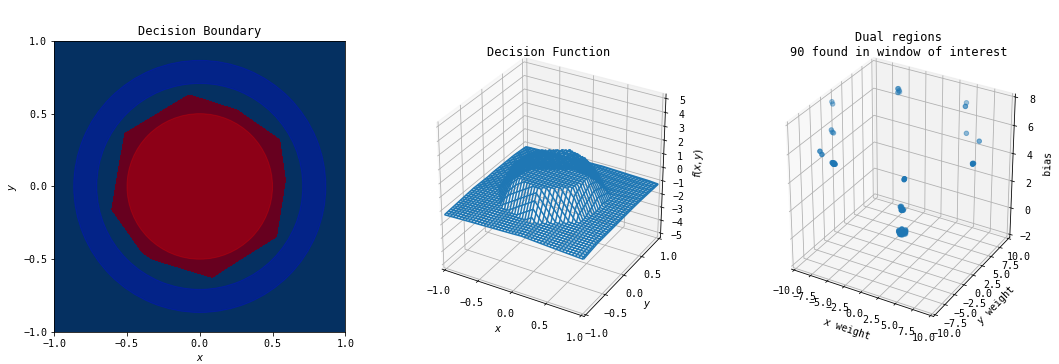}
    
    \includegraphics[width=\textwidth]{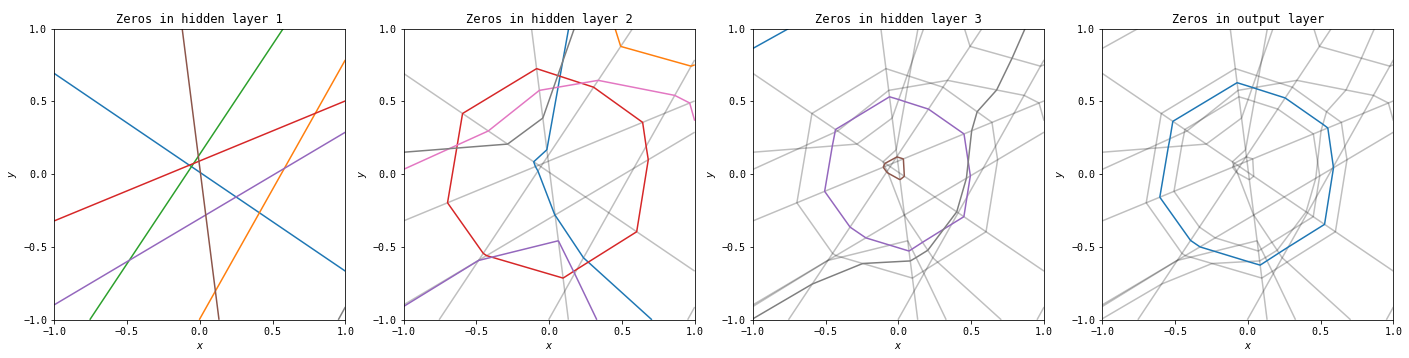}
    
    \caption{The polytopes resulting from the various layers of a simple network to classify a circle versus a surrounding annulus.  
    Top left: the original problem and the decision boundary determined by the network.  Top middle: the outputs of the network.  
    Top right: the dual weights. 
    Bottom first: the zeros for each of the perceptrons from the original input space to the first hidden layer   --- these decision boundaries are all lines, as the perceptrons at this stage are purely linear in the original input.  Each color corresponds to one of eight nodes in this hidden layer (and the colors do not relate between each of the four bottom plots).  Not every node has zeros occuring in the window shown.  
    Bottom second: the zeros for each of the perceptrons from the original input to the second hidden layer, with the boundaries for the first hidden layer in light gray.  These are lines in the output space of the first layer, but appear non-linear when shown in the original input space.  Each boundary can only break at one of the lines from the previous layer.  
    Bottom third: the zeros for each of the perceptrons from the original input to the third layer, with the boundaries of the first two layers in gray.  Breaks in this layer can occur at any location where it crosses a zero of a previous layer.  
    Bottom fourth: zeros in the output layer.  This forms the decision boundary shown in the top left.
    }
    \label{fig:circlesExample}
\end{figure}

One way to think of the polytopes resulting from ReLU activation patterns is the way in which they arise as a consequence of the iterated perceptron structure inherent in this style of network. Each layer builds upon the nonlinearities in the previous layers by drawing a line in the output space of the previous layer.  An example of this is illustrated in Figure~\ref{fig:circlesExample}.  

The first hidden layer of the network, bottom left of Figure~\ref{fig:circlesExample}, is relatively simple --- each of the nodes in the first layer has a line for a decision boundary (where the output of that node switches from positive to negative, resulting in the attenuation by ReLU). Each subsequent layer builds upon the previous. To illustrate this, the decision boundaries highlighted for each layer, in its plot in the bottom row of Figure~\ref{fig:circlesExample}, are reproduced in subsequent layers in gray.  The more complicated decision boundaries for each subsequent layer are always locally linear with changes in direction only arising where they intersect a boundary from a previous layer.  This is a direct result and also illustration of the fact that the nonlinearities of multi-layer networks must be built up from decision boundaries established by the previous layers in the network.  Finally, notice in the bottom right of  Figure~\ref{fig:circlesExample} that the output layer of the network does as expected, constructing a valid piecewise linear approximation to the original classification task. 

\subsection{Extension to Image Data}

\begin{figure}[]
    \renewcommand{\arraystretch}{2}
    \setlength\tabcolsep{1pt}
    \begin{tabular}{C{2cm}|C{1cm}C{1cm}C{1cm}C{1cm}C{1cm}C{1cm}C{1cm}C{1cm}C{1cm}C{1cm}}
    \multirow{2}{*}{\raisebox{-0.5\height}{\includegraphics[width=1cm]{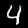}}} & \multicolumn{10}{c}{Output} \\
     & 0 & 1 & 2 & 3 & 4 & 5 & 6 & 7 & 8 & 9 \\\hline
    Dense & 
    \raisebox{-0.5\height}{\includegraphics[width=1cm]{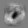}} & 
    \raisebox{-0.5\height}{\includegraphics[width=1cm]{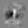}} & 
    \raisebox{-0.5\height}{\includegraphics[width=1cm]{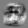}} & 
    \raisebox{-0.5\height}{\includegraphics[width=1cm]{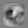}} & 
    \raisebox{-0.5\height}{\includegraphics[width=1cm]{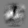}} & 
    \raisebox{-0.5\height}{\includegraphics[width=1cm]{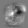}} & 
    \raisebox{-0.5\height}{\includegraphics[width=1cm]{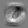}} & 
    \raisebox{-0.5\height}{\includegraphics[width=1cm]{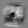}} & 
    \raisebox{-0.5\height}{\includegraphics[width=1cm]{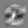}} &
    \raisebox{-0.5\height}{\includegraphics[width=1cm]{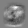}} \\
    Conv & 
    \raisebox{-0.5\height}{\includegraphics[width=1cm]{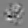}} & 
    \raisebox{-0.5\height}{\includegraphics[width=1cm]{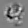}} & 
    \raisebox{-0.5\height}{\includegraphics[width=1cm]{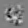}} & 
    \raisebox{-0.5\height}{\includegraphics[width=1cm]{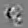}} & 
    \raisebox{-0.5\height}{\includegraphics[width=1cm]{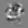}} & 
    \raisebox{-0.5\height}{\includegraphics[width=1cm]{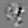}} & 
    \raisebox{-0.5\height}{\includegraphics[width=1cm]{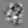}} & 
    \raisebox{-0.5\height}{\includegraphics[width=1cm]{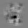}} & 
    \raisebox{-0.5\height}{\includegraphics[width=1cm]{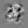}} &
    \raisebox{-0.5\height}{\includegraphics[width=1cm]{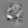}} \\
    Inception & 
    \raisebox{-0.5\height}{\includegraphics[width=1cm]{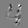}} & 
    \raisebox{-0.5\height}{\includegraphics[width=1cm]{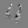}} & 
    \raisebox{-0.5\height}{\includegraphics[width=1cm]{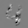}} & 
    \raisebox{-0.5\height}{\includegraphics[width=1cm]{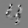}} & 
    \raisebox{-0.5\height}{\includegraphics[width=1cm]{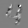}} & 
    \raisebox{-0.5\height}{\includegraphics[width=1cm]{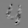}} & 
    \raisebox{-0.5\height}{\includegraphics[width=1cm]{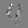}} & 
    \raisebox{-0.5\height}{\includegraphics[width=1cm]{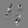}} & 
    \raisebox{-0.5\height}{\includegraphics[width=1cm]{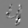}} &
    \raisebox{-0.5\height}{\includegraphics[width=1cm]{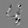}} \\
    ResNet & 
    \raisebox{-0.5\height}{\includegraphics[width=1cm]{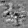}} & 
    \raisebox{-0.5\height}{\includegraphics[width=1cm]{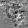}} & 
    \raisebox{-0.5\height}{\includegraphics[width=1cm]{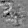}} & 
    \raisebox{-0.5\height}{\includegraphics[width=1cm]{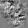}} & 
    \raisebox{-0.5\height}{\includegraphics[width=1cm]{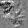}} & 
    \raisebox{-0.5\height}{\includegraphics[width=1cm]{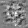}} & 
    \raisebox{-0.5\height}{\includegraphics[width=1cm]{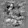}} & 
    \raisebox{-0.5\height}{\includegraphics[width=1cm]{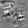}} & 
    \raisebox{-0.5\height}{\includegraphics[width=1cm]{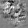}} &
    \raisebox{-0.5\height}{\includegraphics[width=1cm]{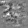}} \\
    \end{tabular}
\caption{The linear regions for each output of the four networks for the input four in the top left corner.  These visualizations are similar to simple forms of saliency mapping~\cite{simonyan2013deep}.}
\label{fig:MNISTRegionsExample}
\end{figure}

The idea of investigating and visualizing linear regions can be extended to higher dimensions and specifically to image data, although visualizations are no longer as simple.  We use the MNIST dataset of handwritten digits which contains 60,000 training samples and 10,000 test samples~\cite{MNIST}.  MNIST was chosen as an image classification dataset due to its relative simplicity.  We used PyTorch~\cite{PyTorch} to train four networks on the MNIST dataset.  These networks are
\begin{itemize}
    \item A dense network with a single hidden layer consisting of 128 nodes.  This network achieves an accuracy of 96.03\%. The training process used cross-entropy loss and PyTorch's SGD function with parameters of 0.01 update rate, 0.5 momentum, 0.01 weight decay, and a batch size of 64 over 30 epochs.
    \item A simple convolutional networks consisting of a convolutional layer with 10 filters and kernel size of 5 followed by a max pool followed by a convolutional layer with 20 filters and a kernel size of 5 followed by a max pool followed by a fully connected layer from 320 nodes to 50 followed by a linear layer from 50 nodes to the 10 outputs.  This network achieves an accuracy of 98.07\%. The training process used cross-entropy loss and PyTorch's SGD function with parameters of 0.01 update rate, 0.5 momentum, 0.01 weight decay, and a batch size of 64 over 30 epochs.
    \item A network with the Inception-v3 architecture as implemented in Torchvision's models subpackage trained from scratch~\cite{torchvision, szegedy2016rethinking}. This network uses more complex layer structures, but to the best of our knowledge none of them result in the network not being a piecewise linear map.  This network achieves an accuracy of 99.08\%.  The first layer of the network was modified to expect images with only one channel and images were upsampled, using bilinear interpolation, to the expected size of 224$\times$224 pixels.  The training process used cross-entropy loss and PyTorch's SGD function with parameters of 0.1 update rate, 0.9 momentum, 1e-4 weight decay, and a batch size of 50 over 22 epochs.
    \item A network with the ResNet-152 architecture as implemented in Torchvision's models subpackage trained from scratch~\cite{torchvision, lin2018resnet}. This network uses more complex layer structures, but to the best of our knowledge none of them result in the network not being a piecewise linear map.  This network achieves an accuracy of 98.92\%.  The first layer of the network was modified to expect images with only one channel.  The training process used cross-entropy loss and PyTorch's SGD function with parameters of 0.1 update rate, 0.9 momentum, 1e-4 weight decay, and a batch size of 50 with 60 epochs.
\end{itemize}

\

For a given input image and a given output node, each network determines a polytope, within the input space, which contains the image. By restricting the neural network to one output node, the gradient of this restricted neural network, at the input image, can be displayed in the same format as the input image. The collection of 40 different gradient "images", computed by considering each of the 4 neural networks and each of the 10 output nodes, are visualized at the given input image in Figure~\ref{fig:MNISTRegionsExample}.  The dense network has relatively little complexity, so it is classifying based on its ``ideal'' shape of each output.  The other networks have more complexity, tend to focus more sharply on the relevant information being passed in, and classify based on that input.  ResNet has behavior that is not as human-interpretable.  The visualization of these linear regions is similar to the idea of saliency mappings, although many modern forms of saliency mapping are more sophisticated than simply visualizing the gradients at an input image, as this is doing~\cite{simonyan2013deep}.

\section{Polytope Evolution Through Training}
\label{sec:animation}

\begin{figure}[]
    
    \hspace{\fill}
    \includegraphics[height=2in]{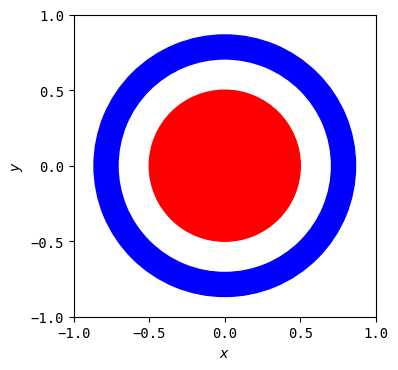}
    \hspace{\fill}
    \includegraphics[height=2in]{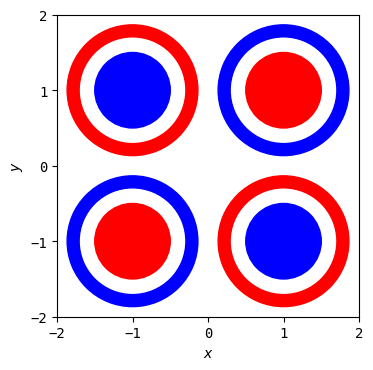}
    \hspace{\fill}
    
    \caption{Two classification problems used to show animations of polytope structures during the training process.  Left: the goal is to classify points in the red annulus as being a separate class as those in the blue annulus.  Right: a a combination of the left problem with the XOR problem to demonstrate more sophisticated network behavior.}
    \label{fig:circleProblem}
\end{figure}

The polytope structures discussed in Section~\ref{sec:polytope-structure} and their associated linear mappings change as the network trains.  For an example of this, we continue with the problem of classifying a circle versus a surrounding annulus and additionally consider a more complex problem that is a combination of the XOR problem and the circle versus annulus problem, both illustrated in Figure~\ref{fig:circleProblem}.

There exist many simple solutions to the single circle versus single annulus problem, but neural networks do not intrinsically take advantage of the rotational symmetry of this problem to express these solutions.  As has been demonstrated previously~\cite{raghu2017expressive, hanin2017approximating}, any network that solves this problem requires a minimum of three hidden nodes in at least one of its hidden layers.  A node in any layer creates a line in the embedding that is its input, but when mapped back to the original input space that line becomes a trajectory that ``breaks'' by re-angling whenever it encounters a line created by the activation boundary of a node in a previous layer.  A network with a maximum width of two is unable to solve this problem as it is unable to create a closed region in the input space.  To see this, note that each layer can only partition space into four regions (both on, one on, the other on, both off), one of which (both off) will be constant.  Due to this, any such network cannot form a closed region in space and will instead have each of its polytopes extend to infinity.

\begin{figure}[]
    \hspace{\fill}
    \includegraphics[width=0.4\textwidth]{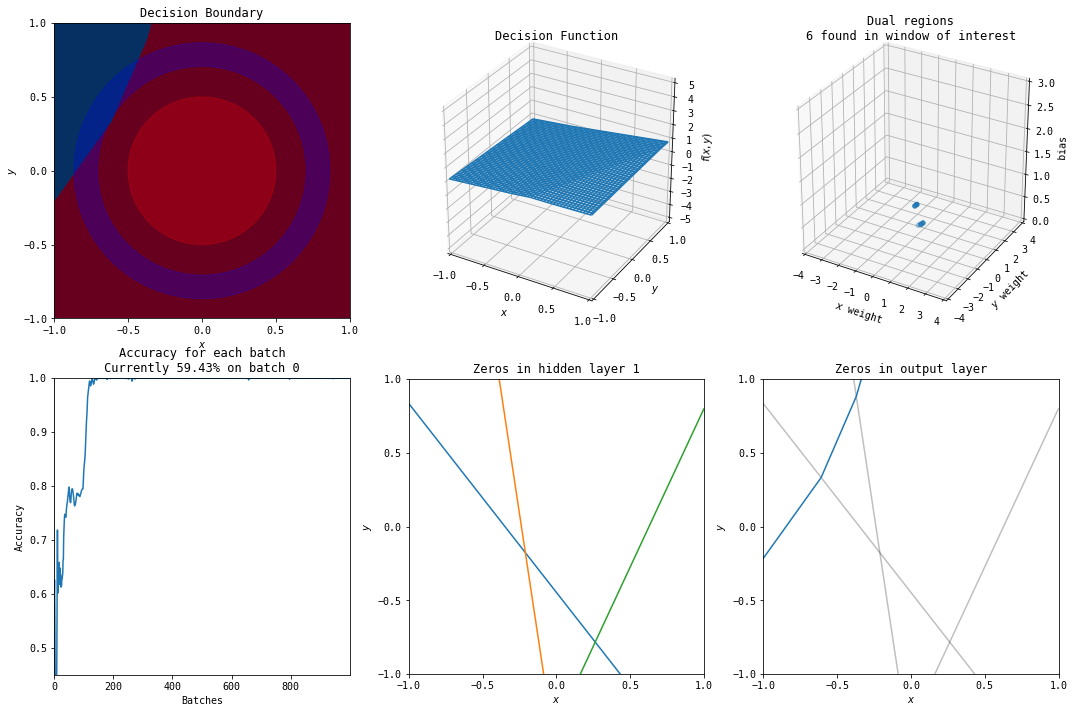}
    \hspace{\fill}
    \includegraphics[width=0.4\textwidth]{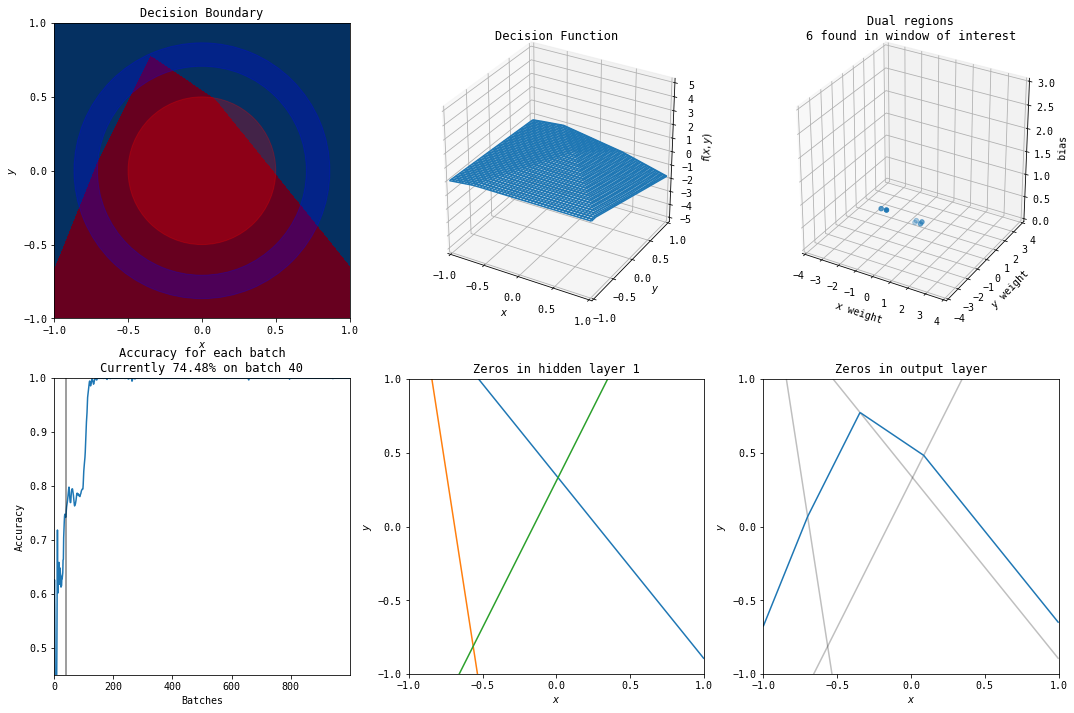}
    \hspace{\fill}
    
    \vspace{1em}
    
    \hspace{\fill}
    \includegraphics[width=0.4\textwidth]{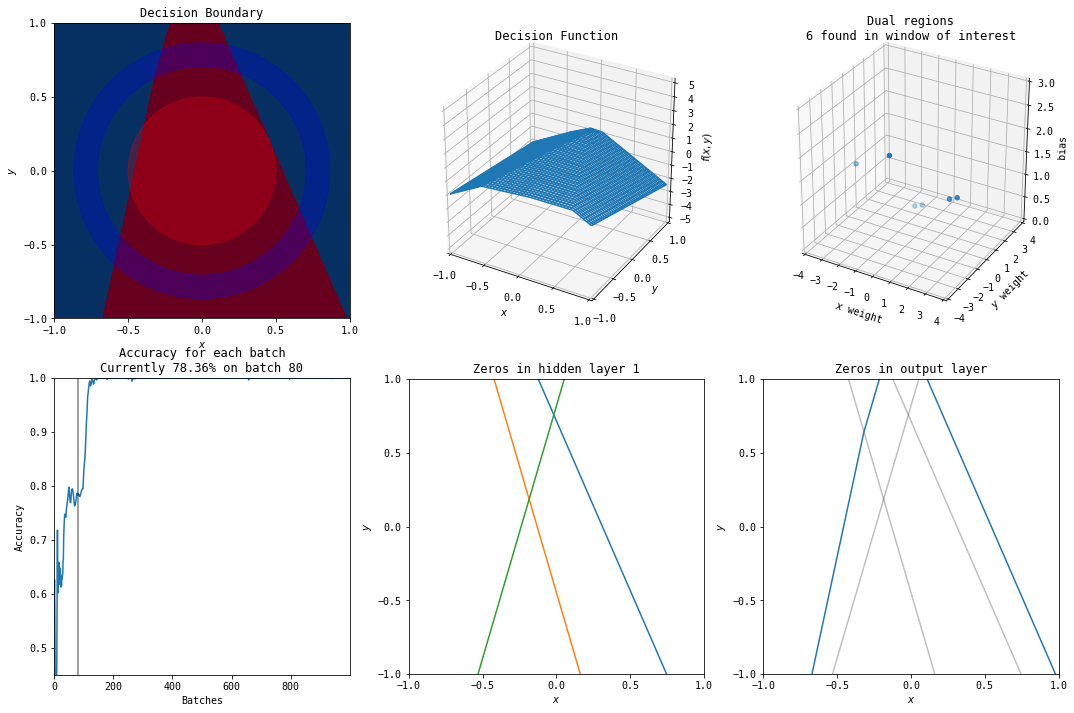}
    \hspace{\fill}
    \includegraphics[width=0.4\textwidth]{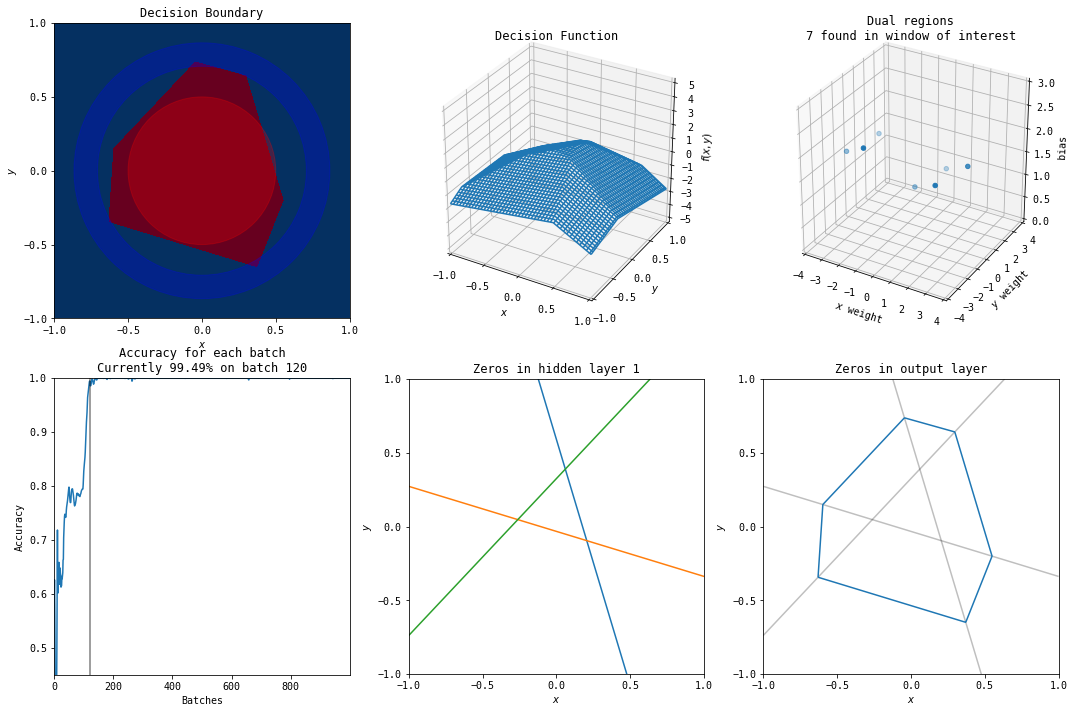}
    \hspace{\fill}
    
    \vspace{1em}
    
    \hspace{\fill}
    \includegraphics[width=0.4\textwidth]{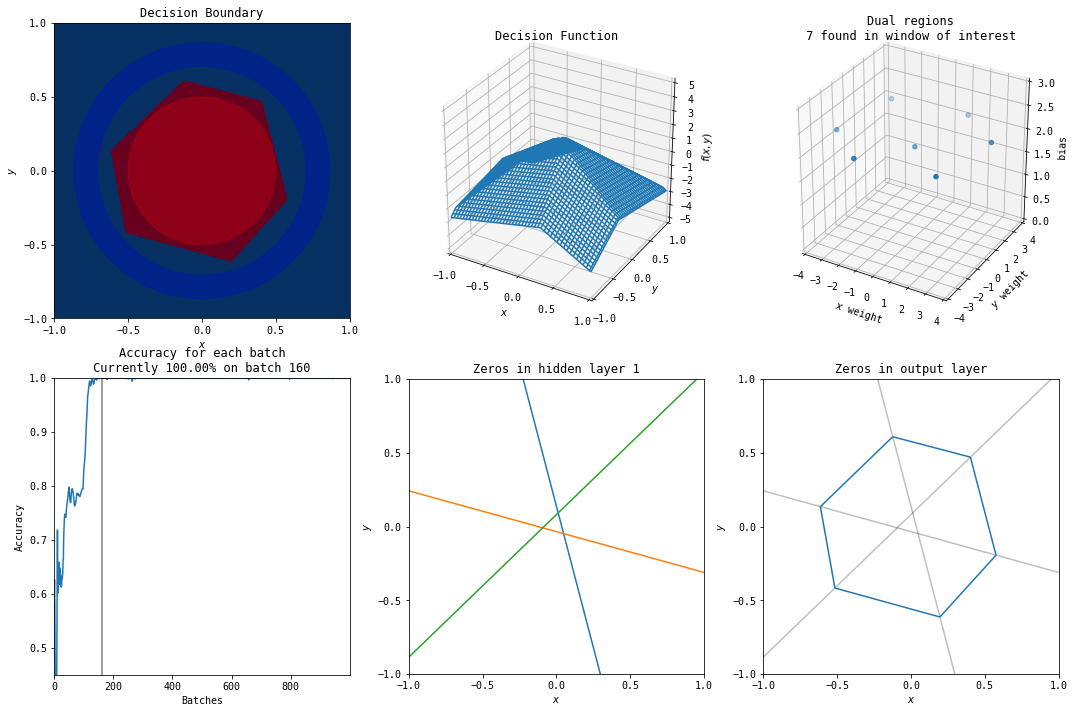}
    \hspace{\fill}
    \includegraphics[width=0.4\textwidth]{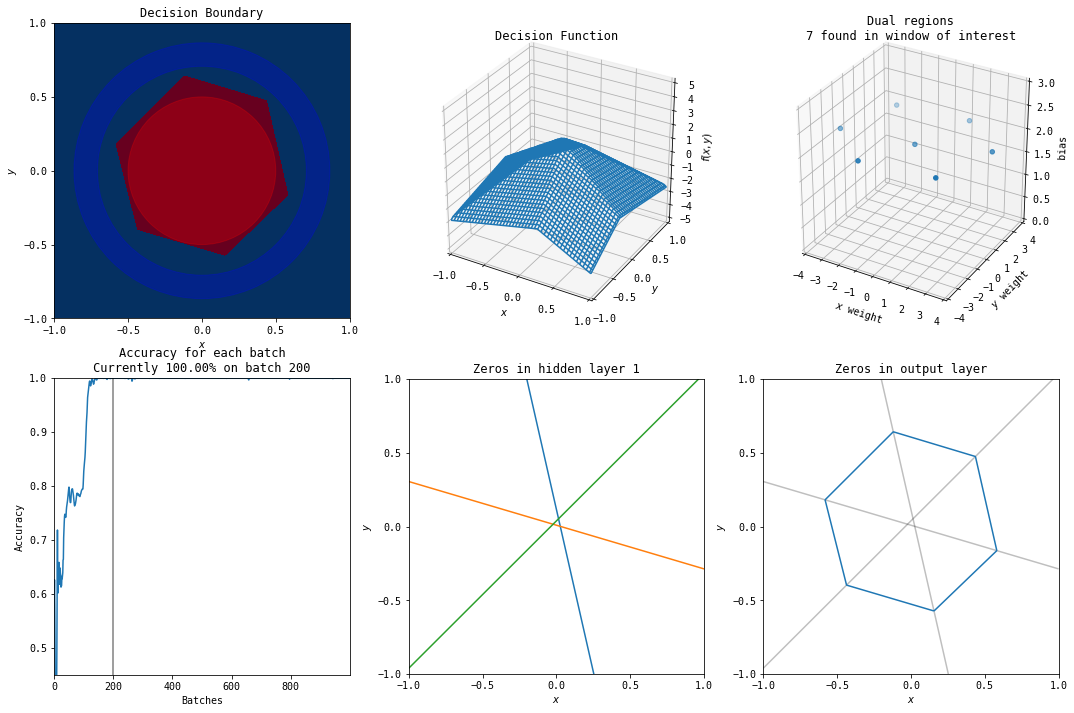}
    \hspace{\fill}

    \caption{The training process of the simplest possible network (three hidden nodes in a single layer) on this problem.  An animation of this process is available at \url{https://www.youtube.com/watch?v=lpXQI-UJIZM}.}
    \label{fig:circleSimpleAnimation}
\end{figure}

\begin{figure}[]
    \centering
    \includegraphics[width=\textwidth]{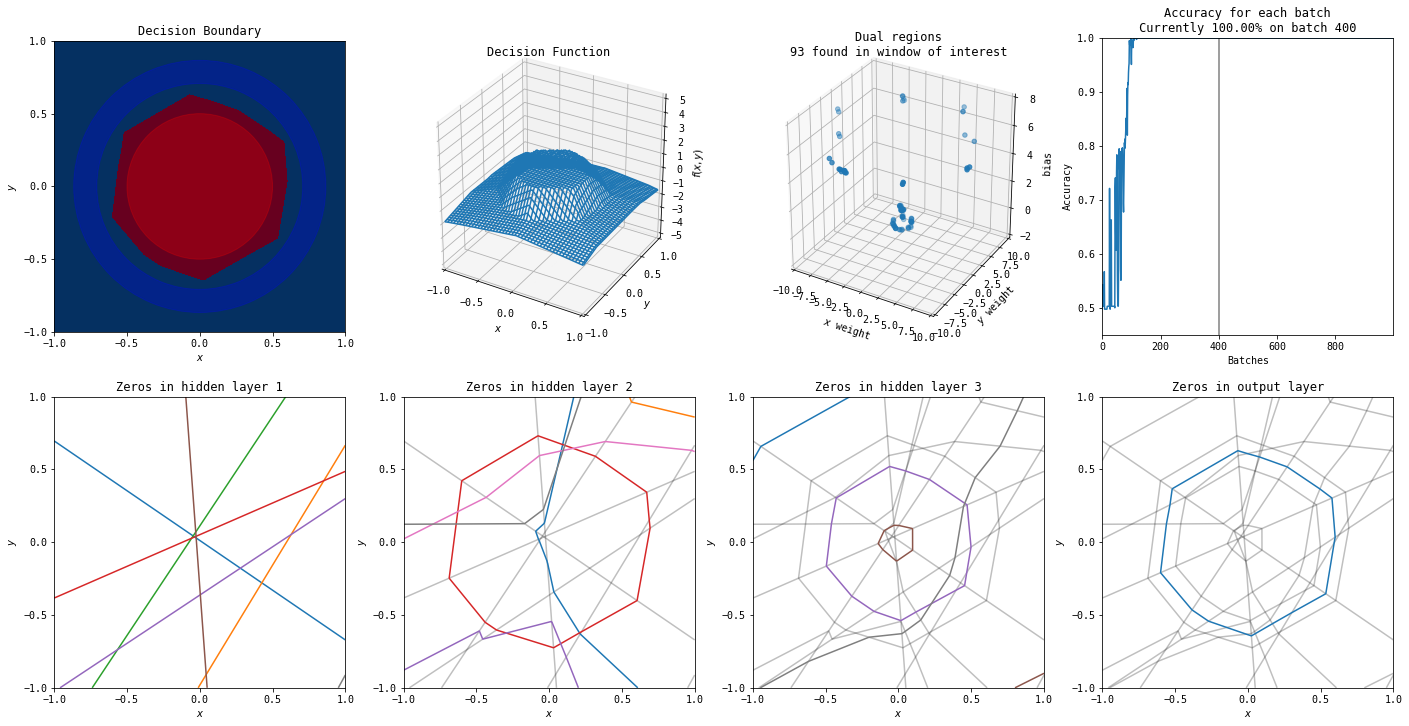}
    \caption{A solution to this problem found by a more complex network (three hidden layers, each with eight nodes).  An animation of the training process of this network is available at \url{https://www.youtube.com/watch?v=rANyD9t-X-c}.}
    \label{fig:circleComplexAnimation}
\end{figure}

To illustrate how these polytopes and decision boundaries change as the neural network trains, we have two examples.  One is the simplest possible network with three nodes in the single hidden layer, and the other is a far more complex network with three hidden layers each containing eight nodes.  Still images of the polytope development throughout the training process for the simple network are shown in Figure~\ref{fig:circleSimpleAnimation} and the end result of the complex network is shown in Figure~\ref{fig:circleComplexAnimation}.  Full videos of the evolution of their polytope structure throughout the training are available at \url{https://www.youtube.com/watch?v=lpXQI-UJIZM} and \url{https://www.youtube.com/watch?v=rANyD9t-X-c}, respectively.

\begin{figure}[]
    \centering
    \includegraphics[width=\textwidth]{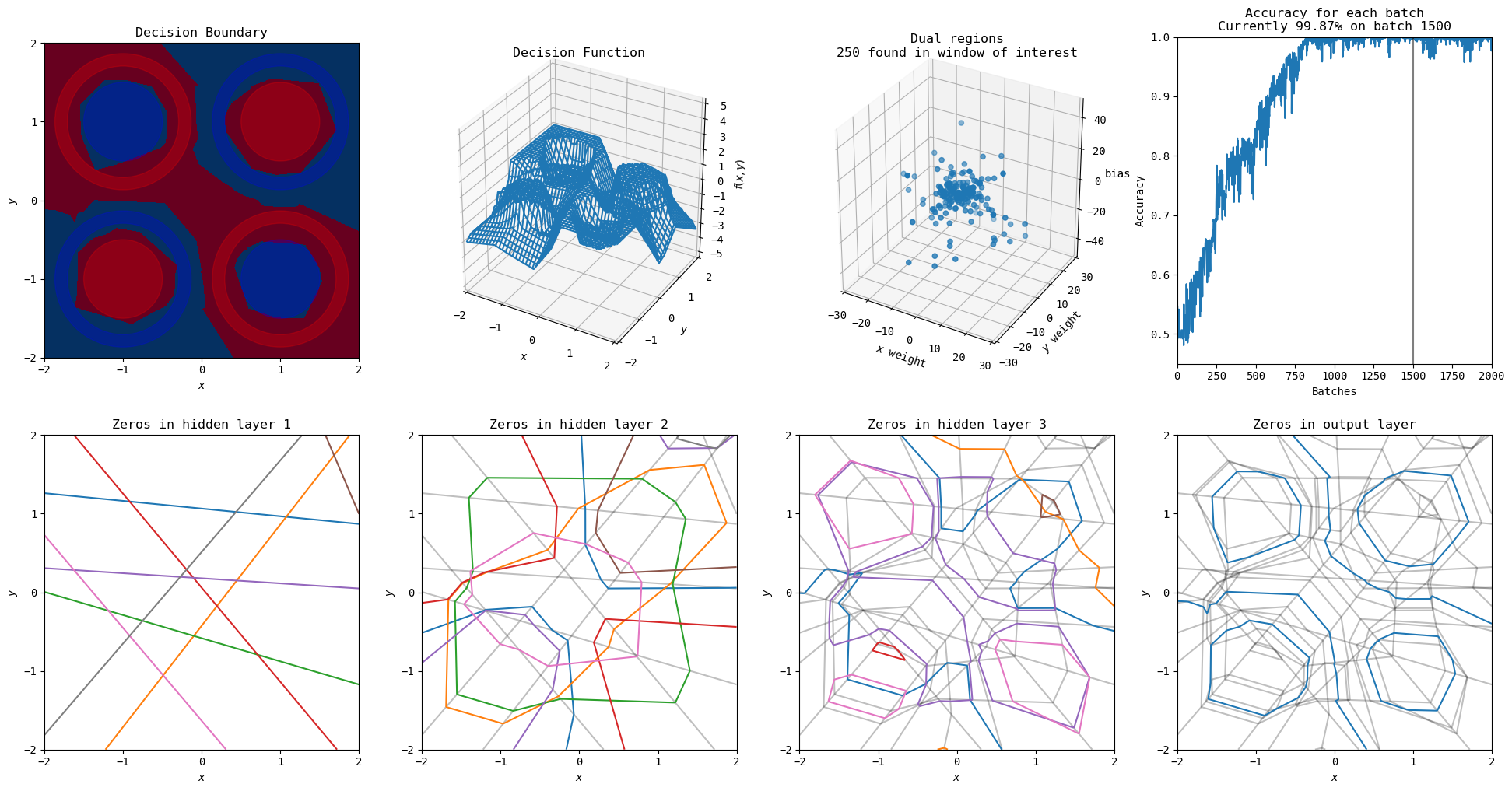}
    \caption{A solution to this problem found by a more complex network (three hidden layers, each with eight nodes).  An animation of the training process of this network is available at \url{https://www.youtube.com/watch?v=T_uoGBUOgUY}.}
    \label{fig:circlesComplexAnimation}
\end{figure}

An example of the polytopes constructed by the more complex network on the more complex problem is in Figure~\ref{fig:circlesComplexAnimation}.  A video of the training process is shown at \url{https://www.youtube.com/watch?v=T_uoGBUOgUY}.

It has previously been shown by Raghu~\etal~\cite{raghu2017expressive} that earlier layers are more important than later layers for the quality of a network and certain visualizations of this were included in their work.  These animations provide additional intuitive examples of this --- the structures constructed by the early layers are passed on, and many of the deeper layers provide only slight modifications to the structures apparent in the first layers.

\section{Region Modifications}

Rather than focusing on the polytope structure of the networks, we can also investigate the affine mappings that arise on each polytope.  This is useful for a number of reasons, but the two simplest are that the visualizations in the previous section cannot be done as simply in high dimension, and that the number of polytopes increases significantly with the complexity of the network.  Even for the simplest network on MNIST, nearly every image in the dataset lies on a unique polytope, and that behavior has been shown to extend to other networks and datasets~\cite{novak2018sensitivity}.  For investigating these affine mappings there are two useful steps to make: constructing notation to allow us to refer to the set of affine mappings potentially used for a specific output of a network, and considering only the affine mappings that are used for training or testing to reduce the number to something computationally manageable.

In terms of notation, the $\matr{W}_{i}$ and $b_{i}$ described in Equation~\ref{eqn:linear-regions} can be written as
\begin{equation}
    \matr{W}_i = \begin{bmatrix} w_{i, 1}^T \\ w_{i, 2}^T \\ \vdots \\ w_{i, o}^T \end{bmatrix} \text{ and } b_i = \begin{bmatrix} b_{i, 1} \\ b_{i, 2} \\ \vdots \\ b_{i, o} \end{bmatrix},
\end{equation}
Where each $w_{i, j}$ and $b_{i, j}$ correspond to the affine mapping in region $i$ for the $j^{\text{th}}$ output of the network.  Then, it is possible to construct the matrix containing the set of linear regions used for a given output, $j$, as
\begin{equation} 
\overline{\matr{C}}_j = \begin{bmatrix} w_{1, j}^T & b_{1, j} \\ w_{2, j}^T & b_{2, j} \\ \vdots \\ w_{m, j}^T & b_{m, j} \end{bmatrix} \in \mathbb{R}^{m \times d+1}.
\end{equation}

In practice, it is computationally infeasible to calculate all $m$ linear regions, so for the purpose of empirical studies we choose $p$ points in the input space to sample and construct the matrix
\begin{equation}
    \matr{C}_j 
    = \begin{bmatrix} w_{1, j}^T & b_{1, j} \\ w_{2, j}^T & b_{2, j} \\ \vdots \\ w_{p, j}^T & b_{p, j} \end{bmatrix}  
    \in \mathbb{R}^{p \times d+1}.
\end{equation}

For simple two-dimensional problems, we choose the $p$ points by sampling from a uniform grid.  We also consider the MNIST dataset~\cite{MNIST}, where the points we sample are the 60,000 training and 10,000 testing input samples from that network.  We construct the $\matr{C}_j$ matrices using the training samples, and we additionally construct $\tilde{\matr{C}}_j$ using the testing samples for evaluation of how various modifications impact accuracy.

\subsection{Clustering Regions}
\label{sec:simplification}

\begin{table}[]
    \centering
    \renewcommand{\arraystretch}{1.5}
    \begin{tabular}{r|rrrrr}
    \# Clusters & Dense & Conv & Inception & ResNet \\\hline
    Original    & 9603  & 9807 & 9908      & 9892 \\
    1           & 8679  & 6766 & 974       & 1432 \\
    10          & 9231  & 8639 & 9660      & 8166 \\
    100         & 9434  & 9382 & 9689      & 8458 \\
    1000        & 9508  & 9586 & 9695      & 8982 \\
    10000       & 9554  & 9696 & 9752      & 9381
    \end{tabular}
    \caption{The accuracies of networks on the MNIST dataset after applying K-means clustering to their collection of local linear maps.  Values reported are the number of correctly labeled test set samples out of 10,000.  Note that the number of clusters for a given network is technically 10 times larger than stated in the table --- for a given output there will be that many clusters, but there are 10 outputs for the MNIST networks.}
    \label{tbl:MNISTClustering}
\end{table}

Even for potentially large numbers of sampled affine maps it is likely that many samples will have a unique $w_{i, j}$ due to the large number of total linear regions.  For example, even simple networks on the MNIST dataset only have overlap on $< 1\%$ of the inputs.  This isn't necessarily surprising, simply due to the sheer number of possible linear regions the network can construct.

However, although these weights are not necessarily equivalent, there is potentially a great deal of redundancy or similarity among them.  As shown in Figure~\ref{fig:XORduals}, patterns appear in the induced linear regions that can indicate redundant behavior.  We can cluster the linear maps and determine how well those clusters are able to replicate the behavior of the network.  One note to make here is that although it would be ideal if we were able to take advantage of this fact to simplify network structures, we don't currently have an algorithm for modifying the neural network based on clustering the points in the dual, so for now it is limited to a tool purely for analysis.  To actually investigate the degree to which this impacts accuracy, we use the MNIST dataset.

The process for evaluating the impact on the MNIST dataset is as follows:
\begin{enumerate}
    \item Calculate the $\matr{C}_j$ and $\tilde{\matr{C}}_j$ matrices.
    \item Train a K-means clustering model on each collection of points in the $\matr{C}_j, j=1,...,10$ matrices.
    \item For each row of each of the $\tilde{\matr{C}}_j$, determine for which cluster center it is closest.
    \item Use that cluster center as a linear mapping from input space to determine the value for that output.
    \item Classify the input based on which of the newly calculated outputs is highest.
\end{enumerate}
Accuracies for this process with different numbers of cluster centers are shown in Table~\ref{tbl:MNISTClustering}. 

Interestingly, the less complex networks capture the linear behavior of the MNIST dataset well.  The dense, single-hidden-layer network in particular is able to recover a solution very close to the best linear classifier in the single cluster case, suggesting it is in some way strongly linear.  This matches previous work that shows that wide networks tend to behave in highly linear ways~\cite{lee2019wide}.

Additionally, although the Inception architecture performs around the accuracy that would be expected from a random classifier with a single cluster, it is able to recover 96.6\% accuracy (better than the original dense network) with as few as 10 clusters per output.

\subsection{Affine Maps Between Sample Centered Local Functions}
\label{sec:mapping}

\begin{figure}[]
    \centering
    \includegraphics[width=0.3\textwidth]{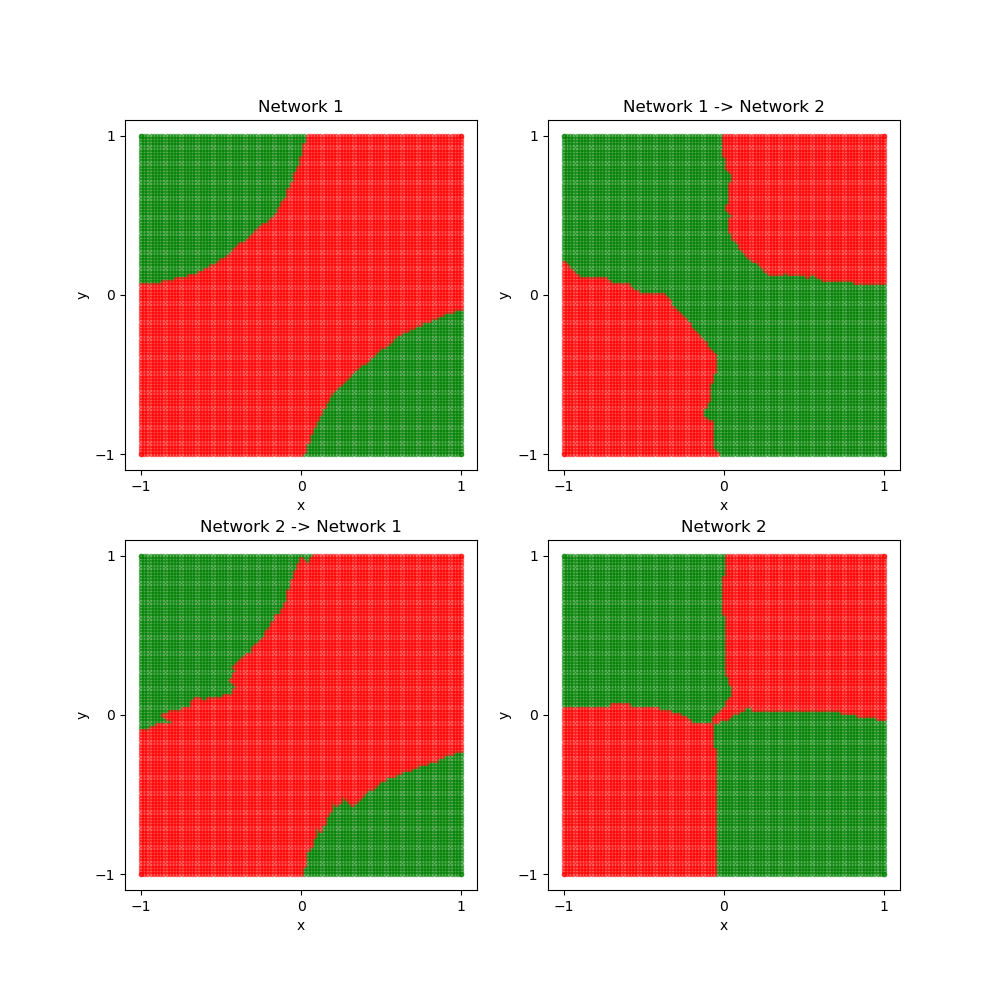}
    \includegraphics[width=0.3\textwidth]{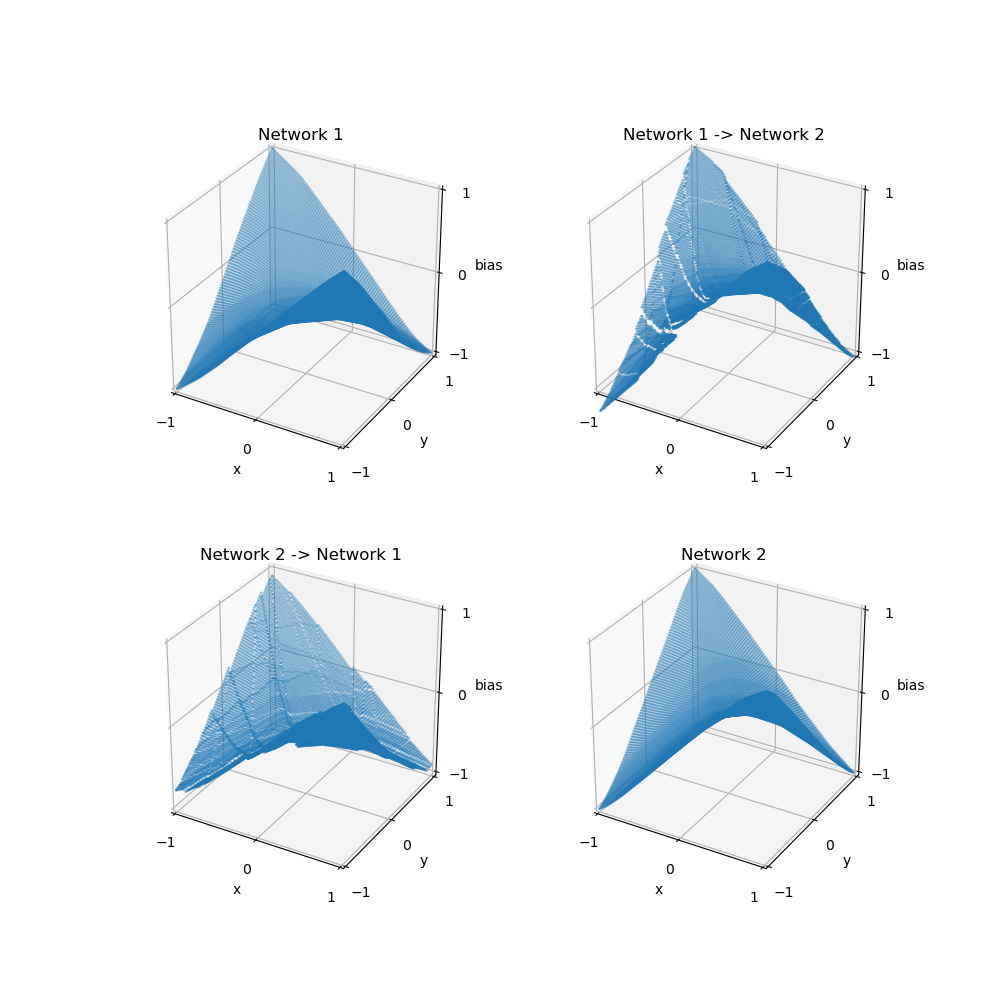}
    \includegraphics[width=0.3\textwidth]{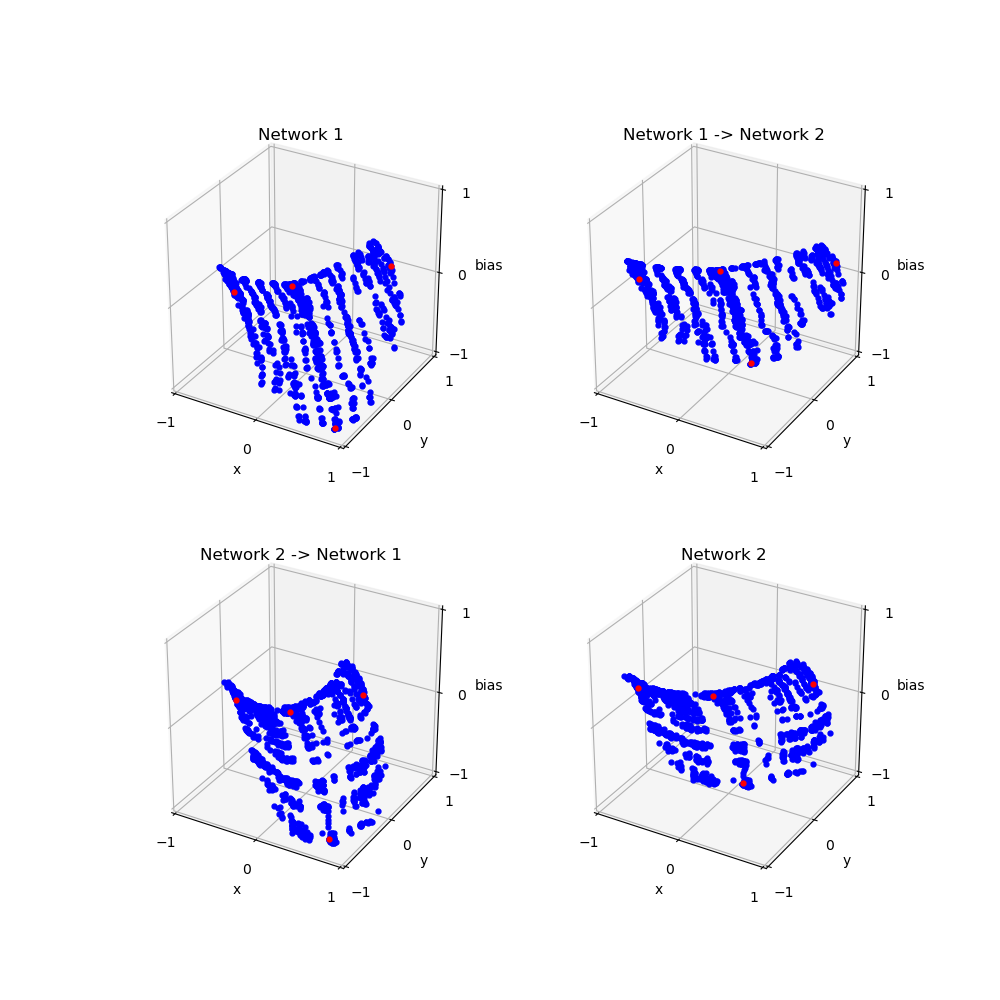}
    \caption{The decision boundaries, wireframes, and weights in the dual space for two different XOR networks and the result of training an affine mapping from the linear regions of one network to the other.  In the dual space the four points of XOR are in red and all other points sampled uniformly from the grid in the original input space are in blue.}
    \label{fig:xorAffineMapping}
\end{figure}

Another area where representing the weights of these linear regions as points in space can be useful is in finding similarities between two networks.  Given $\matr{C}_{j, network 1}$ and $\matr{C}_{j, network 2}$, we can train a least-squares regression model to find a matrix $\matr{M}_j \in \mathbb{R}^{d+1 \times d+1}$ that minimizes
\begin{equation} 
||\matr{C}_{j, network 1} \matr{M}_j - \matr{C}_{j, network 2} ||_2. 
\end{equation}

This method finds a mapping between the linear region weights, or, equivalently, between the gradients of the outputs with respect to the input.  Due to this, as with the K-means clustering method, this method requires running inputs through each original network, calculating the weights, then applying the transformation.

This is similar to the work done by McNeely-White~\etal~\cite{mcneely2019inception} where the authors demonstrated that the outputs of the final layer before the linear classifier of networks trained on ImageNet are affine-equivalent.  Unlike their work, our work investigates the connection between the affine mappings of networks, rather than the feature vectors of networks.

For XOR, we use the $\matr{C}$ matrices (there is only one output) constructed by sampling points on the $101\times 101$ uniform grid.  For MNIST, we use the $\matr{C}_j$ matrices arising from the training set, then evaluate the degree to which the constructed $\matr{M}_j$ reduce accuracy on the $\tilde{\matr{C}}_j$ matrices calculated on the testing set.

Results of this process for the XOR networks are shown in Figure~\ref{fig:xorAffineMapping}.  The resulting points of $\matr{C}_{j, network 1} \matr{M}_j$ are very similar to $\matr{C}_{j, network 2}$ and vice versa.  The function resulting from this is no longer continuous --- because the bias is part of what is being mapped, the result is able to vary based on the position in the plane and regions may no longer join at their boundaries.  

\begin{figure}[]
    \centering
    \tiny
    \bgroup
    \setlength\tabcolsep{1em}
    \begin{tabular}{ccccc}
    Input & Dense & Transform & Prediction & Conv \\[0.5em]
          & $\matr{W}_{0, Dense}$   & $\matr{M}_0$       & $\matr{W}_{0, Dense}*\matr{M}_0$      & $\matr{W}_{0, Conv}$ \\[0.5em]
    & $\mathbb{R}^{60,000 \times (784 + 1)}$ & $\mathbb{R}^{785 \times 785}$ & $\mathbb{R}^{60,000 \times (784 + 1)}$ & $\mathbb{R}^{60,000 \times (784 + 1)}$ \\[0.5em]
    \includegraphics[width=1.5cm]{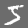} &
    \includegraphics[width=1.5cm]{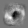} & &
    \includegraphics[width=1.5cm]{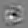} &
    \includegraphics[width=1.5cm]{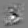} \\
    \includegraphics[width=1.5cm]{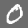} &
    \includegraphics[width=1.5cm]{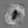} & &
    \includegraphics[width=1.5cm]{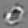} &
    \includegraphics[width=1.5cm]{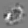} \\
    \includegraphics[width=1.5cm]{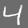} &
    \includegraphics[width=1.5cm]{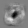} & &
    \includegraphics[width=1.5cm]{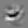} &
    \includegraphics[width=1.5cm]{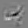} \\
    \includegraphics[width=1.5cm]{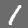} &
    \includegraphics[width=1.5cm]{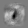} & &
    \includegraphics[width=1.5cm]{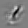} &
    \includegraphics[width=1.5cm]{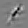} \\
    \includegraphics[width=1.5cm]{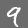} &
    \includegraphics[width=1.5cm]{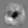} & &
    \includegraphics[width=1.5cm]{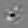} &
    \includegraphics[width=1.5cm]{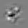} \\
    \end{tabular}
    \egroup
    \caption{An example of the affine mapping from the dense network to the simple convolutional network.}
    \label{fig:MNISTaffineExample}
\end{figure}

This method can also be used to get comparisons on the MNIST dataset.  Examples of this are illustrated for five input samples for $\matr{W}_{0, dense}$ and $\matr{W}_{0, conv}$ in Table~\ref{fig:MNISTaffineExample}.  Qualitatively, the mapped linear regions are similar, but not equivalent, to the target.

\begin{table}[]
    \centering
    \renewcommand{\arraystretch}{1.5}
    \begin{tabular}{rr|rrrrr}
     & &\multicolumn{4}{c}{\textbf{To}} \\
     & & Dense & Conv & Inception & ResNet \\\hline
    \multirow{4}{*}{\rotatebox[origin=c]{90}{\textbf{From}}}
     & Dense     & 9603 & 9536 & 9290 & 9068 \\
     & Conv     & 9567 & 9807 & 9662 & 9588 \\
     & Inception & 8868 & 9488 & 9908 & 9536 \\
     & ResNet    & 9320 & 9738 & 9838 & 9892
    \end{tabular}
    
    \caption{Number of correct labels on the test set (out of 10,000) after applying the affine mapping.  Diagonal elements are the original accuracies of the networks.}
    \label{tbl:affine-mapping}
\end{table}

Table~\ref{tbl:affine-mapping} shows the results of the affine mapping trained on the training set and evaluated on the testing set for the four networks trained on MNIST.  In general, there is degradation in accuracy but a high level of accuracy is nevertheless maintained.  A few interesting points are that Inception does not reproduce the dense network well --- this matches the result that Inception has poor accuracy with one cluster, and suggests that the architecture is doing something that is not as simple.  ResNet is able to reproduce all except the dense network extremely well, suggesting that it forms a strong representation of the data that includes the behavior of the other networks.

One note of caution here is that in some respects this result is not necessarily unexpected --- these four networks are trained to solve the same problem (and even use the same loss function), so on a certain level they are all approximating the same function.  Additionally, the accuracies may not tell the whole story --- although a high level of performance is preserved, it may be the case that the slight variations in accuracy represent significant, qualitatively meaningful differences in what the networks do.  However, these results demonstrate that there is ostensibly an interesting relationship between these different networks and their similar behaviors.  Further research is needed to gauge the extent to which networks trained to solve the problem exhibit equivalent or near-equivalent behavior on a global level.

\section{Conclusion}

We have extended the work of Raghu~\etal~\cite{raghu2017expressive} in visualizing the polytope structure of neural networks with two inputs by constructing animations of the evolution of the polytope structure.  These animations demonstrate how early layers have significant influence over the structure of subsequent layers and how the polytope structures form through training.

Additionally, we have shown experimentally that the number of linear regions that networks have is not necessarily a perfect predictor of their complexity.  The linear regions of all networks considered, except ResNet, can be can be clustered to as few as one or ten cluster centers for networks trained on MNIST while preserving much of their accuracy.

We have also shown experimentally that the linear regions of different networks are similar under an affine mapping.  Applying such an affine mapping preserves a high level of accuracy in the resulting classifier, suggesting that many of the considered networks are solving problems in globally similar ways.

For the future, we are interested in investigating the extent to which these results carry to different datasets and potentially more dissimilar networks --- MNIST is a relatively simple, near linear dataset, and that may potentially skew our results.  Additionally, all four MNIST networks considered were trained using the same process.  Although this demonstrates that their disparate architectures do not seem to differentiate the approximations they learn, investigating to what extent structures such as different loss functions or other forms of regularization impact similarity could prove interesting. 

We also provide support for the tantalizing idea that different networks converge to similar solutions that have a great deal more simplicity than would be suggested by their complex architectures.  We would like to continue to explore the extent to which that idea is correct for modern neural networks.

\bibliographystyle{siamplain}
\bibliography{references}
\end{document}